\documentclass[10pt, a4paper]{article}

\usepackage{lrec-coling2024}
\usepackage{amsmath}
\usepackage{makecell}
\usepackage{amssymb}
\usepackage{multirow}

\title{LHMKE: A Large-scale Holistic Multi-subject Knowledge Evaluation Benchmark for Chinese Large Language Models}

\name{Chuang Liu, Renren Jin, Yuqi Ren, Deyi Xiong\sthanks{~Corresponding author.}}

\address{College of Intelligence and Computing, Tianjin University\\
         Tianjin, China \\
         \{liuc\_09,rrjin,ryq20,dyxiong\}@tju.edu.cn\\}

\abstract{
Chinese Large Language Models (LLMs) have recently demonstrated impressive capabilities across various NLP benchmarks and real-world applications. However, the existing benchmarks for comprehensively evaluating these LLMs are still insufficient, particularly in terms of measuring knowledge that LLMs capture. Current datasets collect questions from Chinese examinations across different subjects and educational levels to address this issue. Yet, these benchmarks primarily focus on objective questions such as multiple-choice questions, leading to a lack of diversity in question types. To tackle this problem, we propose LHMKE, a Large-scale, Holistic, and Multi-subject Knowledge Evaluation benchmark in this paper. LHMKE is designed to provide a comprehensive evaluation of the knowledge acquisition capabilities of Chinese LLMs. It encompasses 10,465 questions across 75 tasks covering 30 subjects, ranging from primary school to professional certification exams. Notably, LHMKE includes both objective and subjective questions, offering a more holistic evaluation of the knowledge level of LLMs. We have assessed 11 Chinese LLMs under the zero-shot setting, which aligns with real examinations, and compared their performance across different subjects. We also conduct an in-depth analysis to check whether GPT-4 can automatically score subjective predictions. Our findings suggest that LHMKE is a challenging and advanced testbed for Chinese LLMs.
 \\ \newline  }

\begin{document}

\maketitleabstract






\section{Introduction}

We have recently witnessed an influx of large language models (LLMs) , which are either proprietary or open-source. Among them, the proliferation of Chinese LLMs \citep{du-etal-2022-glm, DBLP:conf/iclr/ZengLDWL0YXZXTM23, Zeng-arxiv-2021-PanGualpha, Sun-arXiv-2021-ERNIE3.0, Wu-arxiv-2021-Yuan} has been remarkable, with over 60 models unveiled this year alone.\footnote{https://github.com/wgwang/LLMs-In-China} The evaluation of these models has consequently emerged as a critical concern.

\begin{table*}
\centering
\resizebox{\textwidth}{!}{
\begin{tabular}{lcccccc}
\hline
\textbf{Benchmark} & \textbf{Language}& \textbf{\# Tasks} & \textbf{\# Objective Q} & \textbf{\# Subjective Q} & \textbf{\# Numbers ToQ} & \textbf{Standardized S}\\
\hline
    \text{MMCU \citep{DBLP:journals/corr/abs-2304-12986}} & \text{Zh} & \text{51} & \text{11,900} &\text{0} &\text{1} &\sffamily X\\
    \text{C-Eval \citep{DBLP:conf/nips/HuangBZZZSLLZLF23}} & \text{Zh} & \text{52} & \text{13,948} &\text{0} &\text{1} &\sffamily X\\
    \text{CMMLU \citep{DBLP:journals/corr/abs-2306-09212}} & \text{Zh} & \text{67} & \text{11,528} & \text{0}&\text{1} &\sffamily X\\
    \text{M3KE \citep{DBLP:journals/corr/abs-2305-10263}} & \text{Zh} & \text{71} & \text{20,477} &\text{0} & \text{1}&\sffamily X\\
    \text{Xiezhi \citep{DBLP:journals/corr/abs-2306-05783}} & \text{Zh} & \text{516} & \text{249,587} &\text{0} & \text{1}&\sffamily X\\
\hline
    \text{GAOKAO \citep{DBLP:journals/corr/abs-2305-12474}} & \text{Zh} & \text{9} & \text{1,781} &\text{1,030} & \text{3}&\sffamily X\\
    \text{CG-Eval \citep{zeng2023evaluating}} & \text{Zh} & \text{55} & \text{0} &\text{11,000} &\text{3} &\sffamily X\\
\hline
    \text{LHMKE (Ours)} & \text{Zh} & \text{75} & \text{7,884}&\text{2,581} &\text{32}&\text{\checkmark}\\
\hline
\end{tabular}
}
\caption{The comparison between LHMKE and other related benchmarks. Q: Question. ToQ: Type of Question. S: Scoring.}
\label{related}
\end{table*}

Traditional benchmarks may no longer suffice as they are typically designed to assess specific tasks like machine translation or question answering. However, LLMs, having been trained on a variety of instructions \citep{DBLP:journals/corr/abs-2309-15025}, possess the capability to respond to and perform a diverse array of questions and tasks. This indicates a need for more comprehensive benchmarks for their evaluation. A direct approach would be to amalgamate various independent tasks into a unified benchmark for a holistic evaluation of LLMs. Examples of such integrated benchmarks include SuperGLUE \citep{DBLP:conf/nips/WangPNSMHLB19} and BIG-bench \citep{Srivastava-arxiv-2022-Beyond}.



To measure the knowledge acquistion and application of LLMs, simply matching superficial semantic clues in text is not enough. A more effective way to evaluate LLMs is using questions from human exams, which cover various subjects and educational levels. For example, MMLU \citep{DBLP:conf/iclr/HendrycksBBZMSS21} contains 57 subjects from college and high school courses. Some Chinese benchmarks for LLMs, such as M3KE \citep{DBLP:journals/corr/abs-2305-10263}, C-Eval \citep{DBLP:conf/nips/HuangBZZZSLLZLF23}, and CMMLU \citep{DBLP:journals/corr/abs-2306-09212}, follow the same design philosophy as MMLU. However, these benchmarks only focus on one type of questions: multiple-choice questions. This form of questions, despite facilitating the automatic evaluation of LLMs in knowledge application, is not adequate to assess the capabilities of LLMs comprehensively and deeply as LLMs only need to make simple judgments (shortcuts might be exploited during this decision process).

In contrast, human exams include different types of subjective questions, e.g., writing, conditional analysis, conceptual explanations, in addition to multiple-choice questions (MCQs). Different from MCQs, subjective questions are normally equipped with standard or reference answers that are used to compare with answers provided by testers. Because of this, assessing tester answers often requires a broad range of knowledge, rather than word matching. In real human exams, this is usually done by experienced teachers as reviewers. However, this manual assessment is not desirable for testing LLMs as it is usually time-consuming and expensive.

Fortunately, advanced LLMs, such as GPT-4 \citep{OpenAI-OpenAI-2023-GPT-4}, seem to be a promising automatic assessor in comparing answers with reference answers. Recent studies show that advanced LLMs, if equipped with well-designed prompts or personalized roles \citep{DBLP:journals/corr/abs-2308-07201}, are able to compare different pairs of answers and provide scores that are consistent with human evaluators. This inspires and encourages us to build new datasets with multiple types of questions (including subjective questions) for comprehensively and automatically evaluating LLMs.

We hence propose LHMKE, a \textbf{L}arge-scale, \textbf{H}olistic and \textbf{M}ulti-subject \textbf{K}nowledge \textbf{E}valuation benchmark for Chinese LLMs. LHMKE covers 30 subjects with 75 tasks, and each question in LHMKE is sourced from the realistic standard exams with a specific score. This allows us to standardize each subject to a uniform scoring system. We compare LHMKE with other related benchmarks in Table~\ref{related}.

We have evaluated 11 Chinese LLMs on the proposed LHMKE, focusing only on LLMs instruction-tuned by Supervised Fine-tuning (SFT) \citep{Ouyang-arxiv-2022-Training} and Reinforcement Learning from Human Feedback (RLHF) \citep{Christiano-NeurIPS-2017-Deep, Stiennon-arxiv-2020-learning, Ouyang-arxiv-2022-Training} because of their remarkable capabilities \citep{Chung-arxiv-2022-Scaling, Wei-ICLR-2022-Finetuned, Sanh-ICLR-2022-Multitask}. Generally, experimental results show that current Chinese LLMs have difficulty in achieving high scores, with noticeable performance gaps across different subjects. Most LLMs perform better in the elementary and secondary school exams than exams of other education levels. Unsurprisingly, the newest versions of LLMs, such as ChatGLM-6B\footnote{https://github.com/THUDM/ChatGLM-6B} and ChatGLM2-6B\footnote{https://github.com/THUDM/ChatGLM2-6B}, surpass their earlier versions. Moreover, the tested LLMs exhibit expertise in the Teacher Certification  subject within the career development group and the Education subject within the college group. However, the newest versions of LLMs are not always better than old versions. This implies that while current LLMs have improved in subjects related to basic education, they still face challenges in other domains.

Additionally, we have compared various methods for scoring subjective questions in LHMKE. These methods include traditional metrics, using one LLM as a reviewer, and using two LLMs as reviewers. We have also used GPT-3.5\footnote{https://platform.openai.com/docs/guides/gpt} and GPT-4 \citep{OpenAI-OpenAI-2023-GPT-4} as our initial evaluators. We define different prompts to instruct them to mimic human reviewers when grading LLM-autored answers. Our results suggest that GPT-4 with appropromiate prompts matches most closely with human scorers.

\begin{figure*}
    \centering
    \includegraphics[width=0.95\textwidth]{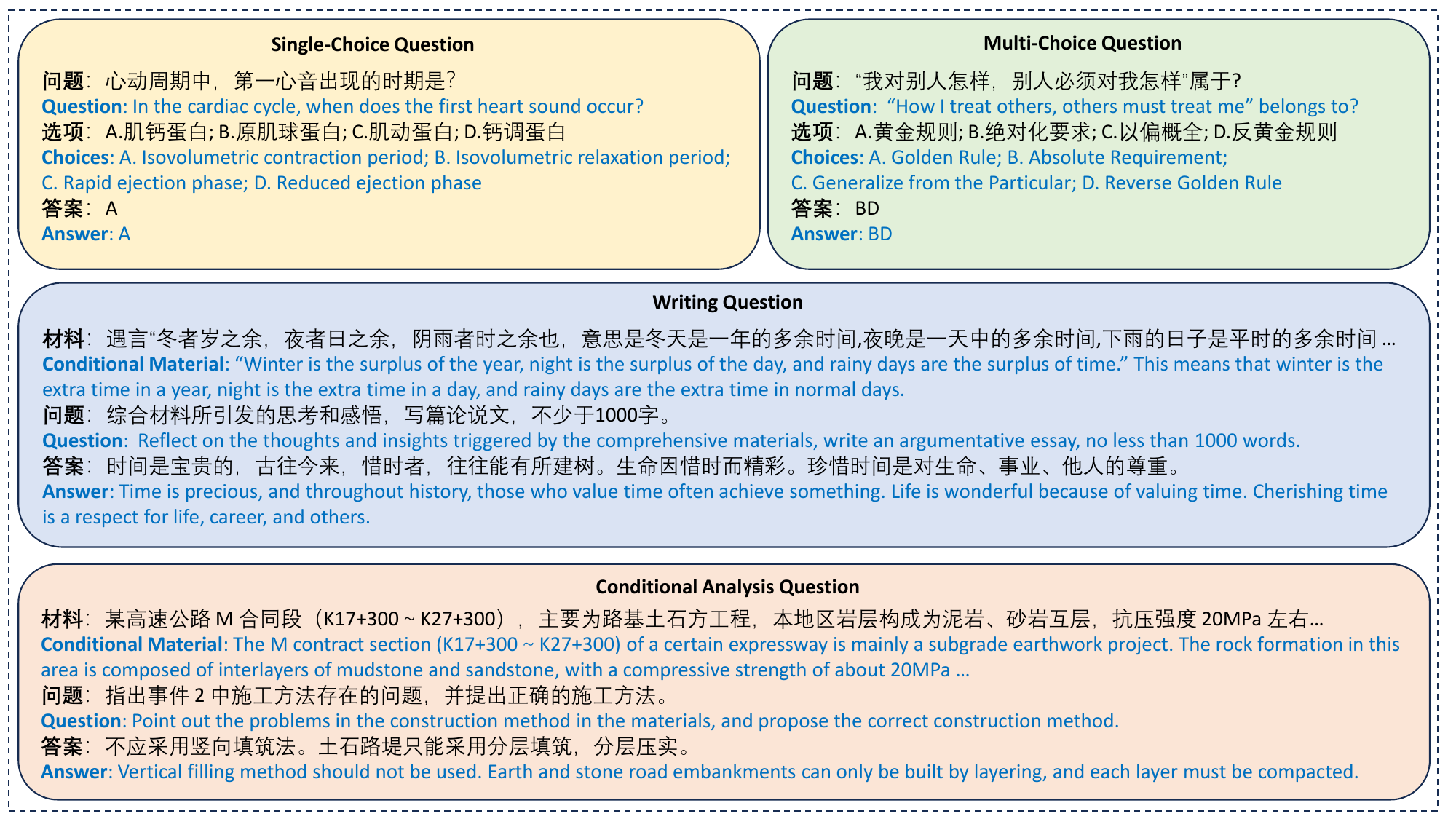}
    \caption{Examples in LHMKE. The yellow example of a objective question with single-choice from Western Medicine subject. The green example of a objective question with multi-choice from Psychological Counselor subject. The blue example of subjective question with writing from Teacher Certification . The orange example of subjective question with conditional analysis from Construction Practical Examination. }
    \label{example}
\end{figure*}

\paragraph{Our main contributions in the paper:}

\begin{itemize}
\item We introduce LHMKE, a comprehensive, multi-subject knowledge evaluation benchmark for Chinese LLMs, which to date encompasses the largest number of question types in alignment with the major Chinese education system.
\item We have conducted tests on a broad of latest open-source SFT/RLHF Chinese LLMs under a zero-shot setting.
\item  We have evaluated the performance of each LLM across different subjects. In addition, various evaluation methods for automatically scoring LLM-generated answers of subjective questions have also been explored on LHMKE. We release LHMKE (data and evaluation scripts) at \url{https://github.com/tjunlp-lab/LHMKE}.
\end{itemize}

\section{Related work}



A variety of benchmarks \citep{DBLP:journals/corr/abs-2310-19736} have been developed to evaluate Chinese LLMs capacity for knowledge acquisition and application. Unlike other datasets designed for assessing language comprehension \citep{xu2023superclue, DBLP:journals/corr/abs-2306-09296}, reasoning \citep{zhang2023exploring, wang2023scibench, DBLP:journals/corr/abs-2312-12853}, role-play \citep{DBLP:journals/corr/abs-2312-16132}, bias \citep{DBLP:journals/corr/abs-2306-16244} and interaction with environments \citep{DBLP:conf/emnlp/LiZ000YLHL23, DBLP:conf/nips/ZhuangYWSZ23, liu2023training, DBLP:journals/corr/abs-2307-13854}, these benchmarks focus on measuring the knowledge acquired during training, which is a fundamental aspect of understanding the capabilities of LLMs. Current Chinese knowledge evaluation benchmarks consist of multiple-choice questions collected from various examinations, with LLM performance evaluated in terms of accuracy.

C-Eval, proposed by \citet{DBLP:conf/nips/HuangBZZZSLLZLF23}, comprises 13,948 multiple-choice questions across 52 tasks. Concurrent to C-Eval, M3KE \citep{DBLP:journals/corr/abs-2305-10263} collects 20,477 multiple-choice questions on 71 tasks. MMCU \citep{DBLP:journals/corr/abs-2304-12986} and CMMLU \citep{DBLP:journals/corr/abs-2306-09212} are similar benchmarks consisting of multiple-choice questions, containing 11,900 and 11,528 questions respectively. Xiezhi \citep{DBLP:journals/corr/abs-2306-05783} stands out due to its size, encompassing 249,587 questions across 516 subjects.

Despite the rapid expansion of Chinese benchmarks, there is a noticeable lack of subjective questions. To enhance the diversity of question types, GAOKAO-Bench \citep{DBLP:journals/corr/abs-2305-12474} and CG-Eval \citep{zeng2023evaluating} are proposed accordingly. GAOKAO-Bench \citep{DBLP:journals/corr/abs-2305-12474} includes real Chinese college entrance examination questions, comprising 1,781 objective questions and 1,030 subjective questions. Meanwhile, CG-Eval \citep{zeng2023evaluating} assesses Chinese text generation capabilities with 11,000 subjective questions across three question types.

In comparison to these works, LHMKE is a comprehensive Chinese benchmark that not only spans the entire Chinese educational spectrum from primary school to career development but also includes both objective and subjective questions from standard Chinese examinations. Figure~\ref{example} shows examples in LHMKE.

\begin{figure*}
    \centering
    \includegraphics[width=\textwidth]{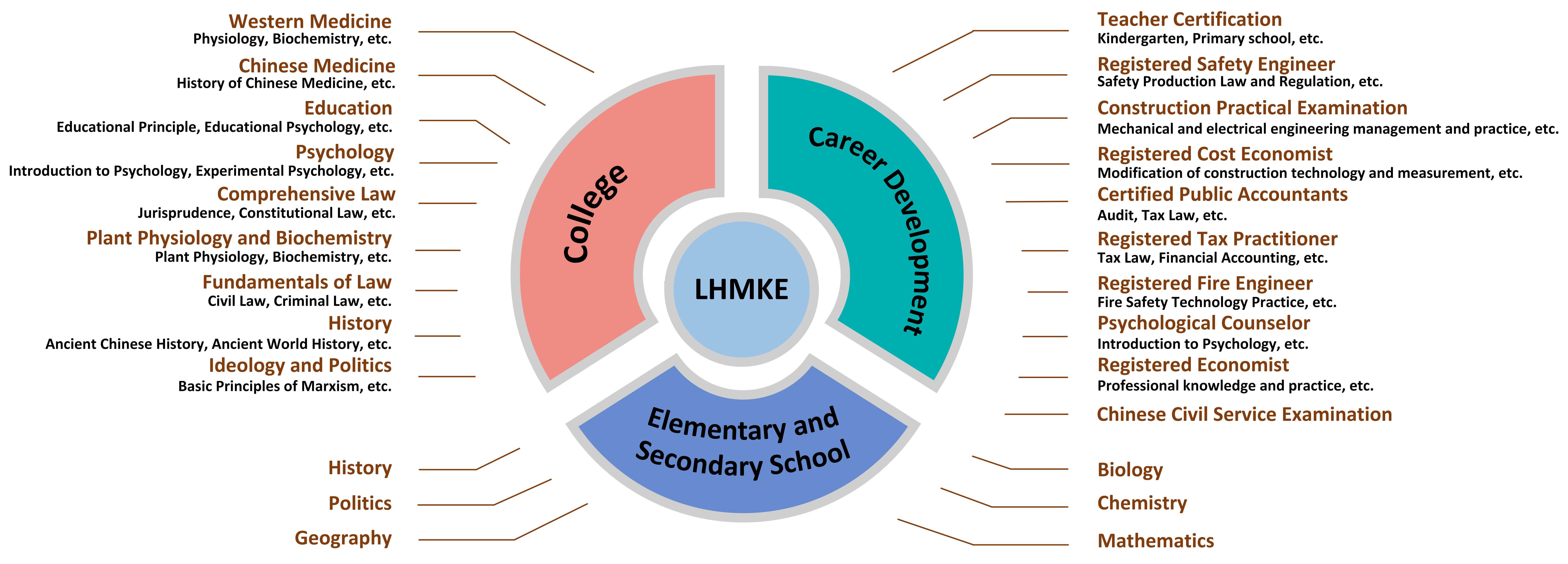}
    \caption{Main subjects in LHMKE.}
    \label{subjects}
\end{figure*}

\section{LHMKE} 

LHMKE encompasses a broad spectrum of Chinese education levels, ranging from primary school tests to professional exams, with a total of 75 tasks across 30 subjects. The data in LHMKE, which are all standard exam questions, have been meticulously collected online by college students to ensure their quality.

LHMKE is divided into three distinct groups: Elementary and Secondary School, College, and Career Development. The elementary and secondary school group includes educational levels of primary school, middle school, and high school. The college group comprises questions from major fields of study in line with the Chinese postgraduate examination. The career development group features questions from popular professional qualification examinations. 

We have instructed collectors to expand the scale of each subject based on its standard score. This means that the total score for each subject should be a multiple of its standard score, facilitating the quantification of LLMs’ performances. Furthermore, we have capped the number of questions in each subject at 300 for efficiency. For example, the score in the primary school math exam is 100 with 20 questions. This indicates that we require 15 sets of primary school math questions to accumulate a total of 300 questions, with the overall scores amounting to 1500. To maintain completeness, we have included several complete examination papers for every subject as far as possible. We only replace undesired questions with those from the same examination papers from other years.

In total, we have amassed 10,465 questions across 30 subjects, which correspond to 75 tasks. All subjects and tasks in LHMKE are presented in Figure~\ref{subjects}.

\subsection{the Group of Elementary and Secondary School}

 This group is composed of nine subjects. We have selected mathematics at the primary school level, as the mathematics subjects at higher educational levels often involve a large number of equations, which is not the primary objective of LHMKE. In addition to mathematics, we also select history, politics, and biology, which are taught in both middle school and high school. Additionally, the subjects of chemistry and geography are included from middle school and high school, respectively.

\subsection{the Group of College}
\begin{table}
\centering
\resizebox{0.46\textwidth}{!}{
\begin{tabular}{lccc}
\hline
 & \textbf{G.ESS} & \textbf{G.C}&\textbf{G.CD}\\
\hline
No.S  &9  & 11&10 \\
No.T & 9 & 37& 29 \\
No.Q  & 3,555 &3,554 & 3,356 \\
No.OQ  & 2,031 &2,819 & 3,034\\
No.SQ  & 1,524 &735 & 322\\
No.Avg.Q  & 395 &323.1 & 335.6 \\
No.Max.Q  & 609 &344 & 421\\
No.Min.Q  & 314 &295 & 270\\
Avg.OQ Tokens   & 127.0 &92.4 &116.2  \\
Avg.OA Tokens  & 1.0 &1.4 & 1.4\\
Avg.SQ Tokens   & 217.7 &189.0 &165.8  \\
Avg.SA Tokens  & 75.0 & 196.7& 74.7\\
\hline
\end{tabular}
}
\caption{\label{statistic} Overall statistics of LHMKE. G.ESS: the Group of Elementary and
Secondary School. G.C: the Group of College. G.CD: the Group of Career Development. S: Subject. T: Task. Q: Question. OQ: Objective Question. SQ: Subjective Question.}
\end{table}

This group consists of 11 subjects with 37 tasks, spanning a variety of fields. Specifically, we include Psychology, Education, History, Ideology and Politics, Western Medicine, Chinese Medicine, Comprehensive Law, Fundamentals of Law, and Plant Physiology and Biochemistry in this group. Each Law subject is divided into sections for law students and non-law students to meet the requirements of Chinese educational departments. Moreover, each subject is a comprehensive examination, in other words, they are all composed of several tasks. For instance, the Psychology subject covers Developmental and Educational Psychology, Experimental Psychology, Introduction to Psychology and Psychology of Teaching.

\begin{table*}
    \centering
    \resizebox{\textwidth}{!}{
    \begin{tabular}{ll}
    \hline
        Components & Prompt\\
        \hline
        Description of the given Role & \makecell[l]{We would like to get your feedback on Answer 2's performance in answering the above user question with\\ reference to the question and Answer 1's answer, as Answer 1's answer is completely correct.\\ Please rate the accuracy of Answer 2's answer to the questions. Both Answer 2 will receive an overall score\\ on a scale of 1 to 10, with higher scores indicating better overall performance.\\ Please first provide a full explanation of your evaluation, avoiding any possible bias, and make sure that\\ Answer 2's score is obtained by referring strictly to Answer 1's response.\\ Then, output two lines representing the scores of Answers 1 and 2. Note that Answer 1's score is always 10.}\\
        \hline
        Scoring Standards & \makecell[l]{Score Answer 2 strictly according to the following scale:\\
    (a) If Answer 2's response answers the question correctly or matches the main points of Answer 1's \\response exactly and does not contain any errors of detail, a score of 10 is given;\\
    (b) If Answer 2's response partially answers the question or partially matches Answer 1's response,\\ a score in the range of 0 to 5 is given;\\
    (c) If Answer 2's answer lacks important details or knowledge points compared to Answer 1's answer \\a score in the range of 0 to 3 is given;\\
    (d) If Answer 2's response is irrelevant to the question or inconsistent with Answer 1's response,\\ a score of 0 is given;\\
    (e) If Answer 2's response has a clear knowledge error, a score of 0 is given;\\
    (f) If Answer 2's response excerpts large portion of content from the question, a score of 0 is given.}\\
    \hline
    \end{tabular}}
    \caption{The prompt given to the evaluator in our experiments.}
    \label{prompt}
\end{table*}

\subsection{the Group of Career Development}
In this group, we have collected 10 Chinese professional qualification examinations as subjects. These include the Chinese Civil Service Examination, Teacher Certification (General Qualifications), Registered Safety Engineer, Certified Public Accountants, Psychological Counselor, Construction Practical Examination, Registered Fire Engineer, Registered Tax Practitioner, Registered Economist and Registered Cost Economist. Similar to the college group, each subject also comprises multiple tasks. For example, the Certified Public Accountants includes four tasks: audit, tax law, corporate strategy and risk management.
\begin{table*}
\centering
\resizebox{\textwidth}{!}{
\begin{tabular}{lcccccccccc}
\hline
\textbf{LLMs} & \textbf{\makecell{P.M\\(100)}} & \textbf{\makecell{M.P\\ (100)}}&\textbf{\makecell{M.H\\(100)}}&\textbf{\makecell{M.B\\ (100)}} & \textbf{\makecell{M.C\\ (100)}}&\textbf{\makecell{H.P\\ (100)}} &\textbf{\makecell{H.H \\(100)}} &\textbf{\makecell{H.B\\ (90)}} &\textbf{\makecell{H.G \\(100)}}& \textbf{\makecell{Totals\\(890)}}\\
\hline
\text{ChatGLM-6B} & 14.4 & 57.6 &56.9 &43 &42.6 &37.4 & 51.9&28.9 &45.1&377.8\\
\text{ChatGLM2-6B} & 44.5& 80.4 & 72.7&73.7 &73.4 & 69.3& 62.2&\textbf{57.8} &59.3&593.3\\
\text{Baichuan2-7B-Chat} & 19.8 &86.7 &79.8 & 64.2&46.0 &74.3 &\textbf{71.8} &47.1 &\textbf{72.7}&562.4\\
\text{Baichuan2-13B-Chat} &\textbf{46.9}  & \textbf{89.8} &83.7 &65.6 & 71.1& \textbf{78.0}& 71.6& 53.6&70.3&630.6\\
\text{BELLE-7B} & 19.2 & 50.7&48.2 & 38.1& 39.0&37.1 & 40.5& 26.7&38.9&338.4\\
\text{Chinese-Alpaca-2-7B} & 16.3 & 50.8 &54.0 &42.1 & 34.1& 30.7& 40.0& 28.1&33.5&329.6\\
\text{Chinese-Alpaca-2-13B} & 12.1 & 62.3& 57.5& 47.8& 38.0&48.2 & 44.7&31.5 &42.0&384.1\\
\text{MOSS-SFT-16B} & 7.3 & 54.1 & 42.6& 29.9& 16.0& 47.1& 37.4& 18.8&55.4&308.6\\
\text{Qwen-7B-Chat} & 41.6 & 86.3& \textbf{85.9}& 75.0& \textbf{76.2}& 73.5& 66.7&55.7 &70.8&\textbf{631.7}\\
\text{InternLM-Chat-7B} & 33.6 & 83.0 &77.3 &\textbf{75.2} &71.1 & 64.6&60.8 &57.7 &52.4&575.7\\
\text{InternLM-Chat-7B-v1.1} & 42.7 & 76.0 &68.8 & 69.8&63.8 &65.9 &61.4 &54.4 &51.9&554.7\\
\hline
\end{tabular}
}
\caption{\label{group 1}Overall results in the Elementary and Secondary School group, and the numbers in the parentheses represent the total score of the subject in the official exam. P.M: Primary School Math. M.P: Middle School Politics. M.H: Middle School History. M.B: Middle School Biology. M.C: Middle School Chemistry. H.P: High School Politics. H.H: High School History. H.B: High School Biology. H.G: High School Geography.}
\end{table*}

\section{Dataset Statistic}
Table~\ref{statistic} presents the overall statistics of LHMKE. The numbers of subjects in three groups are 9, 11 and 10, respectively. And subjects in the elementary and secondary school, college, and career development group involve of different numbers of tasks, individually cover 9, 37 and 29 tasks, respectively. There are 3,555, 3,554 and 3,356 questions in the three group, which can be classified into objective or subjective questions. Specifically, the numbers of objective questions in each group are 2,031, 2,819 and 3,034, and the numbers of subjective question are 1,524, 735 and 322, respectively. Besides, the maximum numbers of questions in the three groups are 609, 344 and 421, and the minimum numbers are 314, 295 and 270, respectively. Additionally, the average lengths of objective questions in each group are 127.0, 92.4, and 116.2 respectively, with corresponding answer lengths of 1.0, 1.4, and 1.4. The average lengths of subjective questions in each group are 217.7, 189.0, and 165.8, with their respective answer lengths being 75.0, 196.7, and 74.7.

\section{Experiments}
We evaluated a series of Chinese LLMs on LHMKE to understand their capabilities on human exams with a wide variety of question types.

\subsection{Assessed LLMs}
We accessed a wide range of Chinese LLMs. These LLMs are instruction-tuned by SFT/RLHF including ChatGLM-6B\footnote{https://github.com/THUDM/ChatGLM-6B}, ChatGLM2-6B\footnote{https://github.com/THUDM/ChatGLM2-6B}, Baichuan2-7B-Chat/13B\footnote{https://github.com/baichuan-inc/Baichuan2} \citep{baichuan2023baichuan2}, Qwen-Chat-7B\footnote{https://github.com/QwenLM/Qwen} \citep{qwen}, MOSS-SFT-16B\footnote{https://huggingface.co/fnlp/moss-moon-003-sft}, BELLE-7B \citep{ji2023exploring}, InternLM-Chat-7B/13B \citep{2023internlm} and Chinese-Alpaca-2-7B/13B \citep{Chinese-LLaMA-Alpaca, taori2023stanford}.


\begin{table*}
\centering
\resizebox{\textwidth}{!}{
\begin{tabular}{lcccccccccccc}
\hline
\textbf{LLMs} & \textbf{\makecell{P\\ (300)}} & \textbf{\makecell{Edu\\ (300)}}&\textbf{\makecell{H\\ (300)}}&\textbf{\makecell{IP\\ (100)}} & \textbf{\makecell{WM\\ (300)}}&\textbf{\makecell{CM\\ (300)}} &\textbf{\makecell{CL\\ (150)}} &\textbf{\makecell{FL\\ (150)}} &\textbf{\makecell{CL$^*$ \\(150)}}&\textbf{\makecell{FL$^*$\\ (150)}}&\textbf{\makecell{PPaB\\ (150)}}& \textbf{\makecell{Totals\\(2350)}}\\
\hline
\text{ChatGLM-6B} & 89.4 & 123.1 & 67.3&56.0 & 98.8&77.0 &61.1 & 34.1&60.4 &41.4 & 26.6&735.2\\
\text{ChatGLM2-6B} & 102.6 & 119.4 &86.9 &60.5 & 78.8&94.3 &67.4 & 45.6& 71.1&53.2 & 39.7&819.5\\
\text{Baichuan2-7B-Chat} & 136.4 &173.5 &173.1 & 59.2&113.0 &88.5 &96.3 &80.9 &95.4 &70.4 &74.7&1161.4 \\
\text{Baichuan2-13B-Chat} & \textbf{158.9} & \textbf{196.4} & \textbf{181.6}& \textbf{73}& \textbf{135.0}& \textbf{103.3}&\textbf{107.9} & \textbf{83.0}&\textbf{104.2} &\textbf{74.3} &\textbf{89.8}&\textbf{1307.4} \\
\text{BELLE-7B} & 78.8 & 83.7& 58.8& 32.0& 83.3&55.5 & 40.0&29.6 &44.0 &30.2 & 31.1&567.0\\
\text{Chinese-Alpaca-2-7B} &65.0  & 91.6 &95.3 & 30.7& 28.5& 21.5&50.0 & 40.4&49.6 &40.3 &44.5&557.4 \\
\text{Chinese-Alpaca-2-13B} & 63.4 &113.2 &95.8 & 38.9& 25.25& 3.0&74.0 & 52.4&67.5 &53.8 & 50.8&638.0\\
\text{MOSS-SFT-16B} & 63.5 & 91.3 & 77.2& 36.8& 39.8& 35.3& 50.8&38.7 &51.9 & 39.6&31.3 &556.2\\
\text{Qwen-7B-Chat} & 107.9 & 144.7& 116.1&54.1 & 114.0&70.8 & 78.7& 70.3&80.6 & 65.3& 56.0&958.5\\
\text{InternLM-Chat-7B} & 95.8 & 110.3 & 83.3& 59.7&112.0 &89.5 &38.9 &65.4 &74.3 & 60.7&71.3&861.2 \\
\text{InternLM-Chat-7B-v1.1} & 122.3 & 130.5 & 102.3&53.1 &114.3 &84.3 & 65.7& 51.2&65.4 &54.9 &45.1 &889.1\\
\hline
\end{tabular}
}
\caption{\label{group 2} Overall results in the College group, and the numbers in the parentheses represent the total score of the subject in the official exam. P: Psychology. Edu: Education. H: History. IP: Ideology and Politics. WM: Western Medicine. CM: Chinese Medicine. CL: Comprehensive Law. FL: Fundamentals of Law. CL$^*$: Comprehensive Law for non-law students. FL$^*$: Fundamentals of Law for non-law students. PPaB: Plant Physiology and Biochemistry.}
\end{table*}

\begin{table*}
\centering
\resizebox{\textwidth}{!}{
\begin{tabular}{lccccccccccc}
\hline
\textbf{LLMs} & \textbf{\makecell{CCSE\\ (100)}} & \textbf{\makecell{TC\\ (150)}}&\textbf{\makecell{RSE\\ (100)}}&\textbf{\makecell{CPA\\ (100)}} & \textbf{\makecell{PC\\ (100)}}&\textbf{\makecell{CPE\\ (120)}} &\textbf{\makecell{RFE\\ (120)}} &\textbf{\makecell{RTP\\ (140)}} &\textbf{\makecell{RE\\ (140)}}&\textbf{\makecell{RCE\\ (100)}}& \textbf{\makecell{Totals\\(1170)}}\\
\hline
\text{ChatGLM-6B} &38.5  & 74.7 & 25.0&23.3 &23.9 &38.8 & 22.7& 19.7&21.0 & 21.5&309.1\\
\text{ChatGLM2-6B} & 28.3 & 102.4 & 30.3& 21.0& 44.9& 34.0& 21.3& 22.4&26.3 & 21.5&352.4\\
\text{Baichuan2-7B-Chat} & 40.9 &108.1 &31.8 &27.8 & 16.3&9.1 &27.7 & 20.5&32.0 &23.0 &337.2\\
\text{Baichuan2-13B-Chat} &39.2  &\textbf{110.8}  & \textbf{38.0}& 29.3& 7.4& \textbf{60.0}& \textbf{31.7}&28.9 & 39.7&27.3&\textbf{412.3} \\
\text{BELLE-7B} & 33.8 & 55.0& 21.5& 14.5&24.2 & 26.1& 23.0& 15.6& 27.0& 17.5&258.2\\
\text{Chinese-Alpaca-2-7B} & 12.0 & 55.7 & 13.0& 11.2& 0.6& 26.1& 14.7&7.04 &5.7 &5.0&151.0 \\
\text{Chinese-Alpaca-2-13B} & 9.2 & 63.7& 11.8&12.1 &1.8 & 35.1& 10.0& 4.8&4.7 &5.5&158.7 \\
\text{MOSS-SFT-16B} & 26.2 & 57.8 & 20.0&12.7 &10.5 & 4.6& 20.0& 11.1&12.7 &12.3&187.9 \\
\text{Qwen-7B-Chat} & 41.3 & 86.1& 32.3&27.7 &30.0 & 54.7& 28.0& 20.4& 34.7&25.5 &380.7\\
\text{InternLM-Chat-7B} &42.9  & 53.7 &35.5 & 20.5&\textbf{47.5} &37.7 &31.3 &36.6 &\textbf{44.7} &\textbf{28.8} &379.2\\
\text{InternLM-Chat-7B-v1.1} &\textbf{43.0}  & 68.5 &31.0 &\textbf{42.8} &43.4 & 47.2&26.0 & \textbf{38.9}&41.7 &25.8 &408.3\\
\hline
\end{tabular}
}
\caption{\label{group 3} Overall results in the Career Development group, and the numbers in the parentheses represent the total score of the subject in the official examination. CCSE: Chinese Civil Service Examination. TC: Teacher Certification (General Qualifications). RSE: Registered Safety Engineer. CPA: Certified Public Accountants. PC: Psychological Counselor. CPE: Construction Practical Examination. RFE: Registered Fire Engineer. RTP: Registered Tax Practitioner. RE: Registered Economist. RCE: Registered Cost Economist.}
\end{table*}
\subsection{Evaluation Metrics}
Due to multiple question types collected in LHMKE, we evaluated the results of LLMs using different metrics. For objective questions, we used accuracy as the evaluation metric.

For subjective questions, drawing inspiration from Chateval \citep{DBLP:journals/corr/abs-2308-07201}, we employed GPT-4 as an evaluator, giving it  the role of a reviewer. As depicted in Table~\ref{prompt}, the prompt is designed with a detailed description of roles and scoring standards.

\subsection{Results}
We compare the overall scores of each LLM across all subject groups, with each group having its own standard score. This makes it convenient to identify the strengths and weaknesses of different LLMs.

Table~\ref{group 1} provides the results of assessed LLMs over the Elementary and Secondary School group. It has been observed that no single LLM outperforms others across all subjects. Qwen-7B-Chat, narrowly leading Baichuan2-Chat-13B, achieves the highest total scores on this group. Baichuan2-Chat-13B achieves the highest scores in 3 subjects, excelling in Primary School Math, Middle School Politics, and High School Politics, which is closely followed by Baichuan2-Chat-7B and Qwen-7B, leading in High School History and High School Geography, Middle School History and Middle School Chemistry, respectively. ChatGLM2-6B and InternLM-Chat-7B are each leading on High School Biology and Middle School Biology, separately. However, other LLMs lag significantly behind these models. When comparing LLMs of different sizes, such as Chinese-Alpaca-2-13B and Chinese-Alpaca-2-7B, the larger model demonstrates superior performance. Similarly, between different versions of the same LLM like ChatGLM-6B and ChatGLM2-6B, the latest version consistently outperforms its predecessor. Yet InternLM-Chat-7B-v1.1 is not as good as InternLM-Chat-7B, which may be caused by the different instruction data used by these two versions. Interestingly, despite MOSS-SFT-16B being the largest model in our experiments, it does not achieve the highest score in any subjects, indicating that model size is not necessarily indicative of performance.

\begin{figure*}
    \centering
    \includegraphics[width=\textwidth, trim=0 230 0 0]{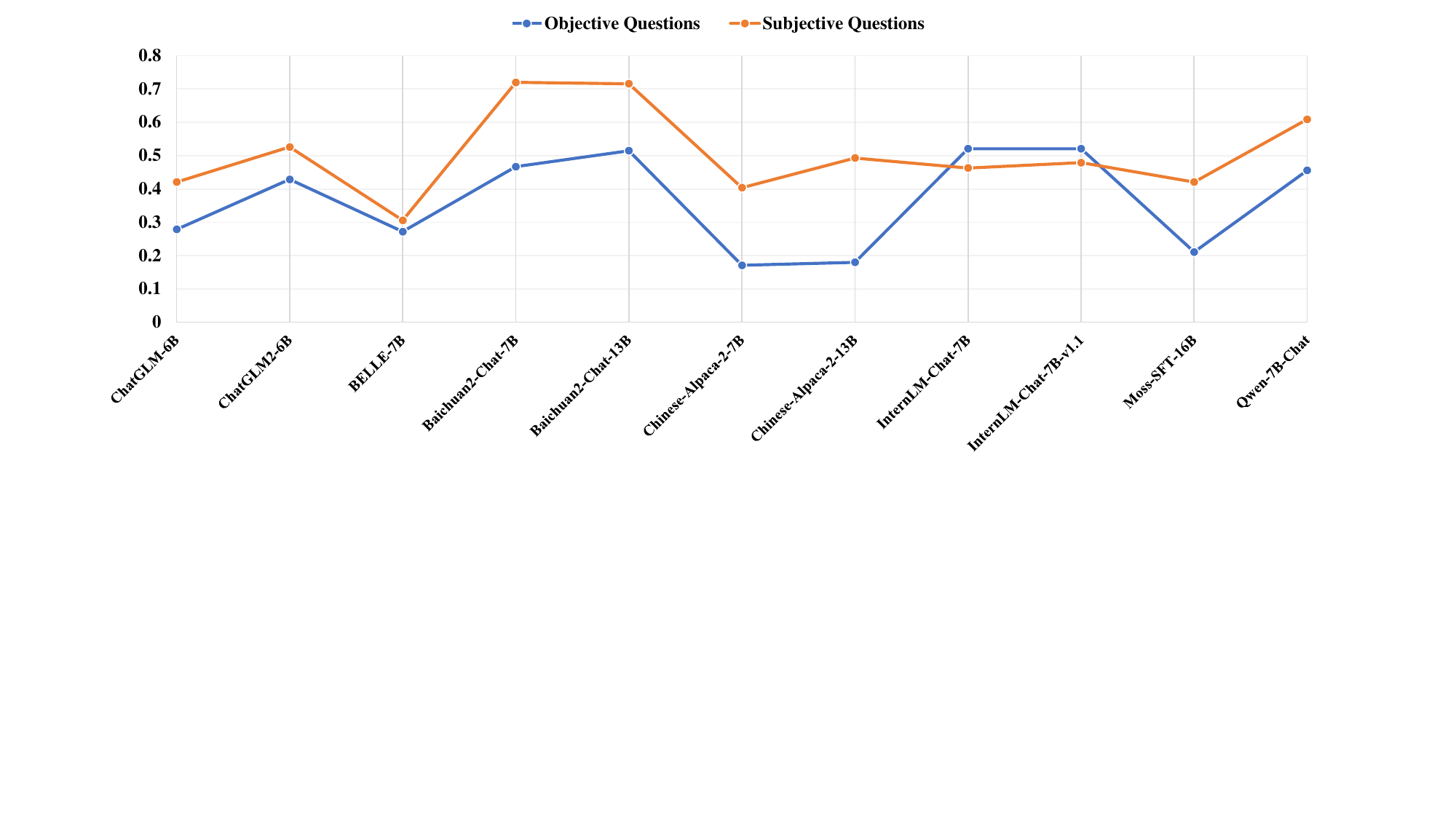}
    \caption{Comparing each LLM's performance in objective questions vs. subjective questions.}
    \label{compared}
\end{figure*}

Unlike the findings in the Elementary and Secondary School group, we observe different trends in the College group (Table~\ref{group 2}). Overall, LLMs tend to perform better in social science subjects than natural science subjects. Most models achieve higher scores in Psychology, Education, and History but struggled with Plant Physiology and Biochemistry and Medicine. On this group, Baichuan2-Chat-13B obtains the highest score in terms of both individual and total scores. Apart from Baichuan2-Chat-13B, Baichuan2-Chat-7B outperforms other models in eight subjects while the remaining highest scores are achieved by ChatGLM2-6B, InternLM-Chat-7B and InternLM-Chat-7B-v1.1, separately. Although Qwen-7B-Chat does not achieve the highest score in any subject, its results are competitive. Interestingly, Chinese-Alpaca-2-13B shows a completely different performance pattern in this group; it outperforms ChatGLM2-6B in History, Comprehensive Law, Fundamentals of Law, Fundamentals of Law for non-law students and Plant Physiology and Biochemistry but scores lowest in Chinese Medicine. This could be attributed to the imbalance in the training data for Chinese-Alpaca-2-13B. Additionally, ChatGLM-6B outperforms ChatGLM2-6B in Education and Western Medicine while Chinese-Alpaca-2-7B follows similar trends to Chinese-Alpaca-2-13B in Psychology and Chinese Medicine.

Finally, we evaluated these LLMs on the Career Development group, as shown in Table~\ref{group 3}. Despite the fact that the top LLM remains Baichuan2-13B-Chat, the overall performance of LLMs in this group is markedly subpar. It is clear that if the subject is closely associated with the educational level in the Elementary and Secondary School group, LLMs are likely to achieve a high score such as Teacher Certification. However, LLMs encounter difficulties with certain professional domain knowledge, such as various types of certification examinations. Furthermore, Baichuan2-13B-Chat has obtained the lowest score in Psychological Counselor compared with its overall performances, which is in stark contrast to its performance of Psychology in the College group. This suggests that even though current LLMs have acquired extensive knowledge from various data sources, a significant gap still exists between them and domain experts.

\section{Analysis}
\label{gpt4}
We provide in-depth analyses of LHMKE, which includes a comparison of each LLM’s overall performance on objective questions versus subjective questions, and an explanation as to why GPT-4 with careful prompting is the most suitable evaluator.

\subsection{Comparing LLM Performance between Objective and Subjective Questions}
Figure~\ref{compared} presents a comparative analysis of the performance of various LLMs on objective and subjective questions. It is clear that the majority of the evaluated LLMs exhibit superior performance on subjective questions as compared to objective questions, with InternLM-Chat-7B and InternLM-Chat-7B-v1.1 being notable exceptions. This, however, should not be misconstrued to suggest that subjective questions are less challenging. In contrast to objective questions, subjective questions offer the possibility of partial scoring even when the answers are not entirely accurate. The most commendable performance on subjective questions is demonstrated by Baichuan2-7B and 13B, while InternLM-Chat-7B and InternLM-Chat-7B-v1.1 emerge as the top-performing LLMs for objective questions. These observations underscore the potential of a balanced mix of objective and subjective questions for a more nuanced evaluation of LLMs.

\subsection{Analysis for Evaluating Subjective Question}
In this section, we conducted a comparative analysis on the evaluation methods for subjective questions, demonstrating that our evaluator outperforms others. Initially, we randomly selected 100 subjective questions from the outputs of evaluated LLMs as examples. Each selection comprised a predicted answer and a corresponding reference answer. Subsequently, we enlisted three postgraduate students to score these predictions, establishing a human benchmark for comparison with our evaluator. This implies that an optimal evaluator would align more closely with human scoring.

To identify the most effective evaluator, we employed GPT-3.5 and GPT-4 as base models. Despite these models having demonstrated their evaluative capabilities, careful design is still required to better align them with human scoring. Broadly speaking, we explored two settings: careful prompting and multi-agent debates. For the careful prompting setting, we designed several prompts, as illustrated in Table~\ref{prompt}, directing the evaluator to adhere to them, thereby enabling precise scoring. For multi-agent debates, we utilized two LLMs as reviewers; after the first reviewer assigned a score, the subsequent reviewer was tasked with verifying the score and granted the authority to modify it if deemed necessary. In addition, we examined major traditional metrics such as BLEU and ROUGE-n.

\begin{table}
\centering
\resizebox{0.5\textwidth}{!}{
\begin{tabular}{llcc}
\hline
\textbf{Metrics}&\textbf{Evaluator} & \textbf{TS}&\textbf{AS}\\
\hline
 \multirow{4}{*}{Traditional Metrics}&BLEU  & 5.0 & 0.05 \\
&ROUGE-1 & 274 & 2.74  \\
&ROUGE-2  & 113 & 1.13 \\
&ROUGE-L & 164 & 1.64  \\
\hline
\multirow{2}{*}{Careful Prompting}&GPT-3.5  &592  & 5.92 \\
&GPT-4 & 500 & 5.0  \\
\hline
\multirow{3}{*}{Multi-agent Debates}&GPT-3.5 \& GPT-3.5  &585  & 5.85 \\
&GPT-3.5 \& GPT-4 & 528 & 5.28  \\
&GPT-4 \& GPT-4 & 503 &  5.03 \\
\hline
Manual Assessment&Human Scorers  & 276 & 2.76\\
\hline
\end{tabular}
}
\caption{\label{debate} Comparing different evaluators with human markers. TS: Total Score. AS: Average Score.}
\end{table}

Table~\ref{debate} presents the outcomes of all experimental evaluators. We observe that all traditional metrics tend to yield a low score, with BLEU being particularly low at 0.05. Although the scores provided by ROUGE are akin to the average human score, they still exhibit less correlation with human scores, as depicted in Table~\ref{coeffient}. When evaluators are given a detailed prompt to guide their scoring, the score assigned by GPT-4 is closer to human evaluators than that of GPT-3.5, with average scores of 5.0 and 5.92 respectively. This suggests that GPT-4 may be a superior evaluator, potentially due to its inherent capability to better understand the question and reference answer compared to GPT-3.5. Results in Table~\ref{coeffient} suggest that the multi-agent debate method, which uses two GPT-4 reviewers achieves the highest correlation to human scorers. Nevertheless, the improvements over a single GPT-4 are marginal and the debate approach would incur double cost. Moreover, while combination of GPT-3.5 and GPT-4 significantly outperforms two GPT-3.5s, it is still inferior to two GPT-4s. Table~\ref{coeffient} also suggests that traditional metrics (e.g., BLEU) are not adequate to evaluate LLMs on subjective questions given their low correlations to human scorers. Consequently, we opted to employ GPT-4 with careful prompting as our experimental evaluator for subjective questions.

\begin{table}
    \centering
    \resizebox{0.5\textwidth}{!}{
    \begin{tabular}{llcc}
         \hline
        \textbf{Metrics}&\textbf{Evaluators} & \textbf{S}&\textbf{P}\\
        \hline
        \multirow{4}{*}{Traditional Metrics}&BLEU  & 0.332 & 0.212 \\
        &ROUGE-1 & 0.398 & 0.347  \\
        &ROUGE-2  & 0.358 & 0.281 \\
        &ROUGE-L & 0.390 & 0.331  \\
        \hline
        \multirow{2}{*}{Careful Prompting}&GPT-3.5  & 0.353 & 0.366 \\
        &GPT-4 & 0.704 & 0.683  \\
        \hline
        \multirow{3}{*}{Multi-agent Debates}&GPT-3.5 \& GPT-3.5  & 0.358 &0.365  \\
        &GPT-3.5 \& GPT-4 & 0.633 & 0.605  \\
        &GPT-4 \& GPT-4 & 0.712 & 0.694  \\
        \hline
    \end{tabular}}
    \caption{Comparing Spearman and Pearson correlation coefficients between different evaluators and human. S: Spearman. P: Pearson.}
    \label{coeffient}
\end{table}

\subsection{Careful Prompting vs. General Prompting}

\begin{table}
    \centering
    \begin{tabular}{lcccc}
    \hline
        \multirow{2}{*}{Indicator} & \multicolumn{2}{c}{GPT-3.5} & \multicolumn{2}{c}{GPT-4}\\
         &GP&CP&GP&CP\\
         \hline
       Average Score & 7.76 & 5.92& 6.83 & \textbf{5.0}\\
       \hline
        Spearman.C  & 0.381 &0.353 & \textbf{0.734} & 0.704\\
        \hline
        Pearson.C  & 0.378 & 0.366& \textbf{0.692}  &0.683\\
        \hline
    \end{tabular}
    \caption{Comparing careful prompting with general prompting. GP: General Prompting. CP: Careful Prompting. C: Coefficients.}
    \label{vs prompts}
\end{table}

Table~\ref{vs prompts} compares the results between careful prompting and general prompting. General prompting refers to the prompts provided to LLMs that do not include the scoring standards shown in Table~\ref{prompt}. Despite general prompting appearing to achieve a higher correlation with human scorers, particularly with GPT-4, it suffers in terms of average score. This suggests that while GPT-4 with general prompting can mimic the process of human scoring, it tends to assign high scores to inaccurate predictions. Therefore, the inclusion of scoring standards in the prompts is necessary.

\subsection{Inter-annotator Agreement Analysis}

We computed the standard deviation of each annotator's scores on 100 randomly selected subjective questions with respect to a reference. Specifically, we observe 23 instances where the standard deviation is 0, 10 instances where the standard deviation falls between 0 and 1, 29 instances where it falls between 1 and 2, 17 instances between 2 and 3, 11 instances between 3 and 4, 6 instances between 4 and 5, and 4 instances where the standard deviation exceeds 5.

We further find that the standard deviation of all annotators on 79 questions is less than 3, indicating that annotators are able to provide similar scores for most questions. Additionally, when a model produces an obviously incorrect response, all annotators provide the same answer. For instance, all annotators assign 0 points to 22 questions simultaneously. However, they collectively award a perfect score of 10 to only one question. This suggests that different annotators maintain distinct scoring criteria to some extent, in line with the nature of subjective evaluation.



\section{Conclusion}
We have constructed a new benchmark, LHMKE, to evaluate Chinese LLMs across a diverse range of question types and subjects, spanning from elementary school to professional certifications. LHMKE includes 30 subjects, 75 tasks, and 10,456 questions. We observe that all evaluated state-of-the-art Chinese LLMs struggle on LHMKE. We will publicly release the benchmark to serve as a new testbed for Chinese LLMs.

\section{Ethics Statement}
This work presents LHMKE, a large-scale, holistic, and multi-subject knowledge evaluation benchmark for Chinese large language models. All data in LHMKE are collected from public sources. All testing instances in LHMKE are carefully scrutinized to exclude any examples with ethical concern. 

\section*{Acknowledgements}
The present research was supported by Zhejiang Lab (No. 2022KH0AB01). We would like to thank the anonymous reviewers for their insightful comments.
\section{Bibliographical References}\label{sec:reference}

\bibliographystyle{lrec-coling2024-natbib}
\bibliography{lrec-coling2024-example}


\end{document}